\documentclass[conference]{IEEEtran}
\IEEEoverridecommandlockouts
\usepackage{cite}
\usepackage{amsmath,amssymb,amsfonts}
\usepackage{algorithm}
\usepackage{algpseudocode}
\usepackage{amsmath}
\usepackage{graphicx}
\usepackage{subcaption}
\usepackage{textcomp}
\usepackage{multicol}
\usepackage{multirow}
\usepackage[dvipsnames]{xcolor}
\usepackage[capitalize]{cleveref}
\usepackage{acronym}
\def\BibTeX{{\rm B\kern-.05em{\sc i\kern-.025em b}\kern-.08em
    T\kern-.1667em\lower.7ex\hbox{E}\kern-.125emX}}

\makeatletter
\newcommand{\linebreakand}{%
  \end{@IEEEauthorhalign}
  \hfill\mbox{}\par
  \mbox{}\hfill\begin{@IEEEauthorhalign}
}
\makeatother

\def\realnumbers{\mathbb{R}}
\def\complexnumbers{\mathbb{C}}

\begin{document}
\acrodef{AI}[AI]{Artificial Intelligence}
\acrodef{SSM}[SSM]{State-space model}
\acrodef{RNN}[RNN]{Recurrent Neural Network}
\acrodef{QAT}[QAT]{Quantization-aware Training}
\acrodef{PTQ}[PTQ]{Post-training Quantization}
\acrodef{GPU}[GPU]{Graphical Processing Unit}

\title{Quantizing Small-Scale State-Space Models for Edge AI\\

\thanks{* equal contribution as first author; § equal contribution as last author}
}

\author{\IEEEauthorblockN{Leo Zhao\textsuperscript{ *}}
\IEEEauthorblockA{\textit{Institute of Neuroinformatics} \\
\textit{University of Zürich and ETH Zürich}\\
lezhao@ethz.ch\\
Zürich, CH \\ }
\and
\IEEEauthorblockN{Tristan Torchet\textsuperscript{ *}}
\IEEEauthorblockA{\textit{Institute of Neuroinformatics} \\
\textit{University of Zürich and ETH Zürich}\\
tristan.torchet@uzh.ch\\
Zürich, CH \\ }
\linebreakand
\IEEEauthorblockN{Melika Payvand}
\IEEEauthorblockA{\textit{Institute of Neuroinformatics} \\
\textit{University of Zürich and ETH Zürich}\\
melika@ini.uzh.ch\\
Zürich, CH \\
}
\and
\IEEEauthorblockN{Laura Kriener\textsuperscript{ §}}
\IEEEauthorblockA{\textit{Institute of Neuroinformatics} \\
\textit{University of Zürich and ETH Zürich}\\
laura.kriener@uzh.ch\\
Zürich, CH \\ }
\and
\IEEEauthorblockN{Filippo Moro\textsuperscript{ §}}
\IEEEauthorblockA{\textit{Institute of Neuroinformatics} \\
\textit{University of Zürich and ETH Zürich}\\
filippo.moro@uzh.ch\\
Zürich, CH \\ }
}

\maketitle

\begin{abstract}
State-space models (SSMs) have recently gained attention in deep learning for their ability to efficiently model long-range dependencies, making them promising candidates for edge-AI applications.
In this paper, we analyze the effects of quantization on small-scale SSMs with a focus on reducing memory and computational costs while maintaining task performance.
Using the S4D architecture, we first investigate post-training quantization (PTQ) and show that the state matrix $\mathcal{A}$ and internal state $\textit{x}$ are particularly sensitive to quantization.
Furthermore, we analyze the impact of different quantization techniques applied to the parameters and activations in S4D architecture. To address the observed performance drop after \ac{PTQ}, we apply \ac{QAT}, significantly improving performance from 40\% (PTQ) to 96\% on the sequential MNIST benchmark at 8-bit precision.
We further demonstrate the potential of \ac{QAT} in enabling sub 8-bit precisions and evaluate different parameterization schemes for \ac{QAT} stability.
Additionally, we propose a heterogeneous quantization strategy that assigns different precision levels to model components, reducing the overall memory footprint by a factor of 6x without sacrificing performance.
Our results provide actionable insights for deploying quantized SSMs in resource-constrained environments.
\end{abstract}


\section{Introduction}
\acp{SSM}~\cite{gu2020hippo,gu2111efficiently,gu2021combining} have recently garnered significant attention as highly effective sequence-modeling architectures. 
\acp{SSM} offer a principled framework for capturing long-range dependencies through structured linear dynamical systems, acting as promising alternatives for \acp{RNN}, which struggle to achieve the same performance, and attention-based models, which scale unfavorably in computational complexity. 
Recent advances, such as the Structured State-space Sequence model (S4) \cite{gu2111efficiently} and its more efficient variant S4D \cite{gu2022parameterization}, have demonstrated strong empirical results on a variety of sequence tasks while maintaining favorable computational properties, including linear time and space complexity during training. 
For inference, these models can be run in the recurrent form with linear time and constant space complexity. These advantages render \acp{SSM} particularly attractive for edge-AI applications with strict constraints on latency, memory, and power consumption. 
By combining modeling capacity with efficient inference, \acp{SSM} emerge as a promising class of models for real-time, on-device processing of temporal data such as speech \cite{schone2024scalable, miyazaki2024exploring, 10448152}, bio-medical signals \cite{tran2024eeg, li2024harmamba}, and sensor streams.

\begin{figure*}[t]
    \centering
    \includegraphics[width=0.92\textwidth]{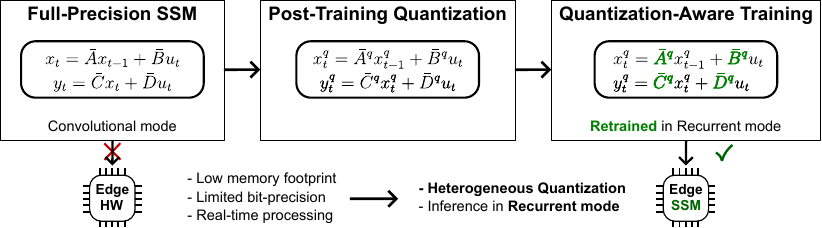}
    \caption{Quantization of small-scale \ac{SSM} networks for edge \ac{AI}. Edge \ac{AI} processors impose constraints on the SSM models, such as limited memory footprint and limited bit-precision arithmetic. We adapt the pre-trained convolutional-mode SSM to these constraints first with \ac{PTQ}, showing that this incurs a consistent performance drop. We then recover the lost performance with recurrent-mode \ac{QAT} and a heterogeneous quantization of the \ac{SSM} parameters that minimizes the memory footprint while maximizing performance.}
    \label{fig:F1}
\end{figure*}

Despite their parameter efficiency, deploying \acp{SSM} on edge devices remains a challenge due to limited available hardware resources in such environments~\cite{suwannaphong2025optimising} (\cref{fig:F1}, left).
Methods to reduce the computational complexity of \acp{SSM} include sparsifying intra-layer communication by introducing binary (\textit{Heaviside}) activation functions~\cite{stan2024learning, karilanova2024zero}, which can be viewed as a \textit{thresholding} operation or \textit{spiking} non-linearity in a Spiking Neural Network. 
We focus instead on quantization, which has emerged as a key technique to address these constraints by reducing the bit-width of model weights and activations~\cite{ghamari2021quantization,novac2021quantization}. 
Quantization not only reduces memory usage but also enables low-precision arithmetic, which simplifies computations and reduces energy consumption \cite{jacob2018quantization}.
However, the effectiveness of quantization varies widely across model architectures and applications. 
In models like \acp{SSM}, which maintain and update internal states over time, aggressive quantization can impair stability and degrade overall performance~\cite{abreu2024q}. 
Previous works investigated the quantization of large-scale \acp{SSM} for efficient deployment on \acp{GPU}~\cite{chiang2024quamba, li2025qmamba, pierro2024mamba}, using only \ac{PTQ} techniques. 
Another work analyzed the performance of a quantized S5 \ac{SSM} model \cite{abreu2024q}, while \cite{meyer2024diagonal} implemented an \ac{SSM} network on the Loihi neuromorphic chip \cite{orchard2021efficient}.

To the best of our knowledge, here for the first time we comprehensively study the effects of both \ac{PTQ} and \ac{QAT} on small-scale, low-precision \acp{SSM}. 
Understanding how different parts of the model tolerate quantization and how to design robust quantization schemes is essential for enabling reliable and efficient edge deployment. Furthermore, existing works do not address the effects of quantization in recurrent mode, which is essential for deployment in resource-constrained hardware.

In this work, we present a systematic study of quantization using fundamental techniques in a small-scale recurrent S4D trained on the sequential MNIST (sMNIST) task, a standard benchmark for sequence modeling.

\begin{figure*}[t!]
    \centering

    \begin{subfigure}[b]{0.38\textwidth}
        \centering
        \includegraphics[width=\textwidth]{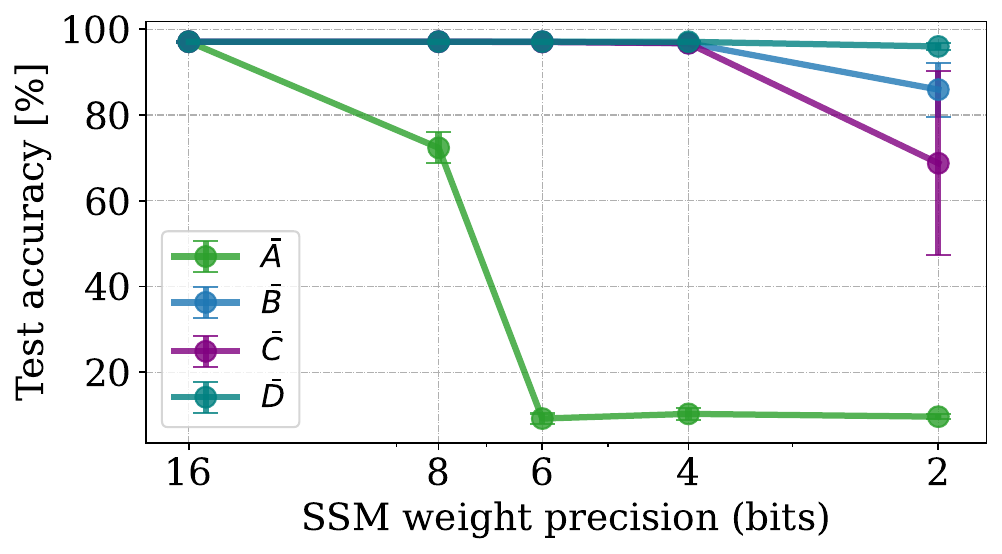}
        \caption{}
        \label{fig:ptq-a}
    \end{subfigure}
    \hspace{0.75cm}
    \begin{subfigure}[b]{0.38\textwidth}
        \centering
        \includegraphics[width=\textwidth]        {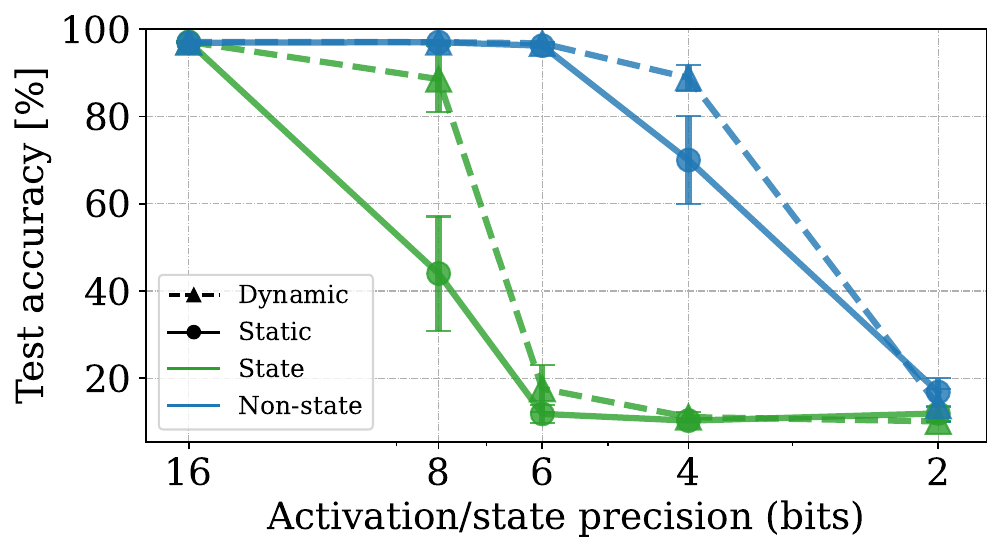}
        \caption{}
        \label{fig:ptq-b}
    \end{subfigure}

    \vskip\baselineskip \vspace{-0.5cm}

    \begin{subfigure}[b]{0.38\textwidth}
        \centering
        \includegraphics[width=\textwidth]{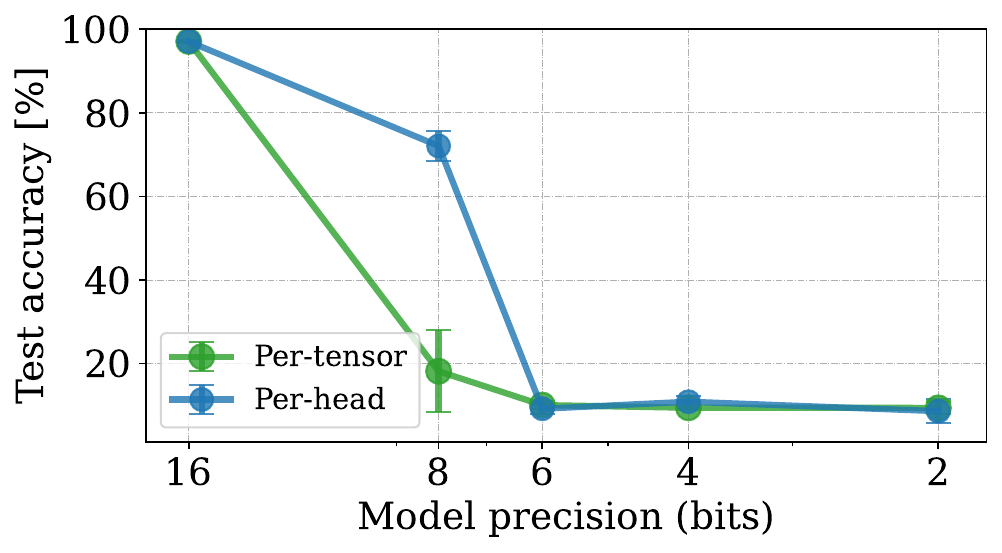}
        \caption{}
        \label{fig:ptq-c}
    \end{subfigure}
    \hspace{0.75cm}
    \begin{subfigure}[b]{0.38\textwidth}
        \centering
        \includegraphics[width=\textwidth]{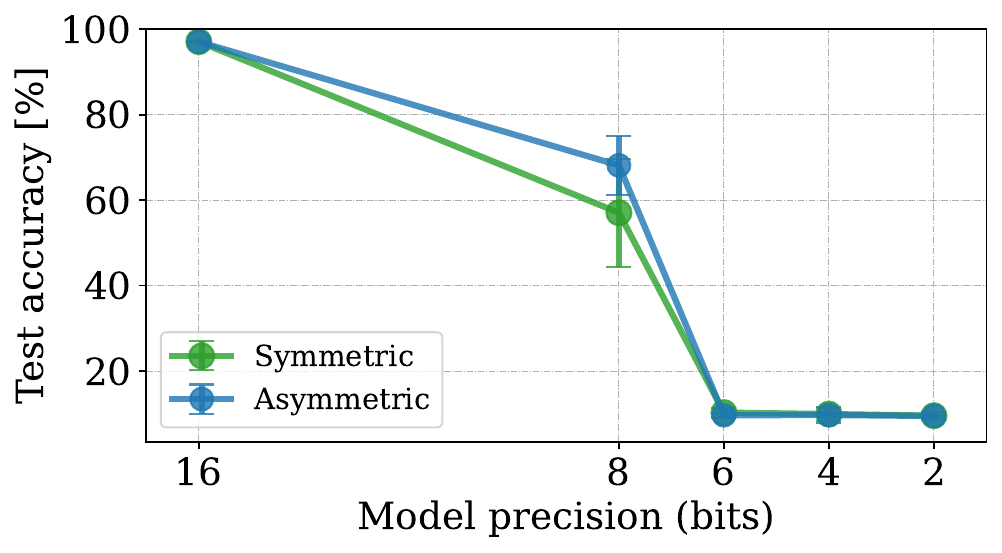}
        \caption{}
        \label{fig:ptq-d}
    \end{subfigure}
    \caption{Accuracy on sMNIST using \ac{PTQ} of different S4D model components with 16 heads and different quantization modes. (a) Quantization of a single state-space matrix (\(\bar{A}\), \(\bar{B}\), \(\bar{C}\), or \(\bar{D}\)). (b) Quantization of state or all other activations (non-state) for \textit{static} and \textit{dynamic} calibration schemes. (c) Effect of quantization granularity at given model bit precision. (d) Effect of \textit{symmetric} and \textit{asymmetric} quantization at given model bit precision. All experiments are performed on 5 different seeds, showing the standard deviations as error bars.}
    \label{fig:F2_PTQ}
\end{figure*}

\subsection{State-space Models for Edge AI}
Several variations of SSMs have been developed lately \cite{gu2020hippo, gu2021combining, gu2111efficiently}, offering different compromises between computational efficiency and performance. These models are built around the state-space model originating in control theory that describes continuous-time linear time-invariant (LTI) systems, which can be expressed by the following equations:
\begin{equation} \label{eq:LTI_eqns}
	\begin{split}
		\dot{x}(t) &= Ax(t) + Bu(t) \\
		y(t) &= Cx(t) + Du(t)
	\end{split}
\end{equation}
where $x(t) \in \complexnumbers^N$ is the latent state, $u \in \realnumbers$ is the input and $y \in \realnumbers$ is the output. $A \in \complexnumbers^{N\times N}, B\in \realnumbers^{N\times1}, C \in \realnumbers^{1\times N}, D \in \realnumbers^{1\times1}$ are the \textit{state-space matrices}. $N$ is the state size. We call $A, B, C, D$ the \textit{state transition matrix}, the \textit{input matrix}, the \textit{output matrix} and the \textit{feedthrough matrix}, respectively. 

In practice, $u$ and $y$ are discrete input and output sequences, and the time-continuous form of the SSM must be discretized during inference, resulting in the equations seen in \ref{fig:F1}. The discrete forms of the state-space matrices are denoted with a bar, e.g. $\bar{A}$. Typical discretization methods are the zero-order hold (ZOH) or bilinear method.

In the S4 architecture, the continuous-time state-space matrices are trained along with the time step $\Delta$, then discretized during the forward pass. The state-space matrices in S4D are initialized according to the HiPPO framework \cite{gu2020hippo}, which compresses past input history into a fixed-size hidden state using orthogonal projections designed to optimally preserve recent information \cite{gu2022train}. 
In particular, the $\bar{A}$ matrix in S4D is simplified to the diagonal form for computational efficiency. 
The SSM layer consists of $H$ parallel and independent SSMs, where $H$ is the number of heads. The outputs of each SSM are passed through a non-linearity and mixed to produce the layer output. 
The depth of the network consists of $D$ stacked SSMs, where $D$ is the depth.
\begin{algorithm}
\caption{S4D Architecture with $D$ Layers and $H$ Heads}
\label{algo:S4D_summary}
\textbf{Input:} Sequence $u^0 = \{u_1, u_2, \dots, u_L\}$\\
\textbf{for} $l = 1$ to $D$ \textbf{do} \hspace{1em} \% Repeat SSM layers\\
\hspace*{1em}\textbf{for} $h = 1$ to $H$ \textbf{do} \hspace{1em} \% Multi-head parallelism\\
\hspace*{2em}Compute $y_l^h = \text{SSMhead}(u^l)$\\
\hspace*{1em}\textbf{end for}\\
\hspace*{1em}Concatenate heads: $y^l = \text{Concat}(y_l^1, \dots, y_l^H)$\\
\hspace*{1em}Apply non-linearity: $y^l = \text{GeLU}(y^l)$\\
\hspace*{1em}Update input: $u^{l+1} = \text{GLU}_l( \text{Conv1D}_l(y^l) )$\\
\textbf{end for}\\
\textbf{Output:} $out = \text{Linear}(u^D)$
\end{algorithm}

A summary of the S4D architecture is provided in \cref{algo:S4D_summary}.
\begin{table}[h]
	\centering
        \caption{Full precision baseline results on sMNIST.}
	\begin{tabular}{p{1.35cm}|p{1.35cm}|p{2.0cm}|p{1.35cm}}
		Model & \#Params & Test accuracy[\%] & StdDev[\%] \\
		\hline
		\textbf{S4D-16h} & 3,850 & 97.200 & 0.250 \\
		\textbf{S4D-64h} & 21,514 & 98.792 & 0.119 \\
		\textbf{S4D-256h} & 184,330 & 99.248 & 0.066 \\
		\textbf{S4D-512h} & 630,794 & 99.270 & 0.061 \\
		\hline
	\end{tabular} \\[5pt]
	\label{tab:fp-results}
\end{table} 



\section{Method} \label{sec:method}

The S4D offers two modes of execution: the \textit{convolutional} and \textit{recurrent} modes. The first exploits parallel computation in \acp{GPU}, but requires processing the entire input sequence in parallel, a requirement that likely is not satisfied in an edge case where inputs are directly streamed from a sensory system. Instead, the recurrent mode treats each input data point sequentially, and is thus compatible with edge processing for inference. The propagation of quantization errors or noise in the recurrence during execution at the edge is also more accurately modeled by the recurrent form of the model. For comparison, \cite{abreu2024q} quantizes the S5 model run with the \textit{associative scan} technique. This technique computes the \ac{SSM} output in parallel and ignores the recurrently propagated error of the quantized state \textit{x} computed in recurrent mode. We are thus interested in investigating the effect of quantization on the model in recurrent mode.

We first pre-train our convolutional S4D models on the sequential MNIST task (sMNIST) in full precision and obtain the baseline classification accuracies summarized in~\cref{tab:fp-results}.
For all model sizes, we train for 30 epochs with a learning rate of $1\mathrm{e}{-3}$, following the training procedure described in the S4D paper \cite{gu2022parameterization}.
In convolutional mode, the SSM output $y$ is computed through the convolution $y = \bar{K}*u$, where the convolutional kernel $\bar{K} = \sum^n_i C\bar{A}^i \bar{B}$ is efficiently computed exploiting the diagonality of $\bar{A}$. 
We then switch the model to its recurrent form and proceed with the quantization of the S4D parameters for \ac{PTQ}. 

For the SSM, we directly quantize the discrete state-space matrices. As $\bar{A}$, $\bar{B}$, and $\bar{C}$ are complex-valued tensors, we separate the real and imaginary parts but always quantize both parts according to the same scheme to reduce the complexity of cross-multiplications in future hardware implementations.
Moreover, quantizing the state \textit{x}, which is calculated recurrently, is necessary in recurrent mode to take advantage of limited bit-precision in the SSM parameters. Our experiments show that the performance of the quantized S4D models is extremely sensitive to the precision and stability of the state \textit{x}. As such, we heuristically clip the real and imaginary components of the state to the interval $[-50, +50]$ to prevent divergence over time. Activations outside the SSM are real-valued. The SSM layer also includes a GeLU non-linearity, which we quantize as $\text{qGELU}_q(x) = x \cdot \min(\max(0, x+2), 1)$, following the method developed in \cite{abreu2024q}. 

Quantization can be employed with an enormous selection of techniques. We thus select the most fundamental, commonly used quantization parameters for our experiments. We map the full-precision range to the quantized range uniformly, using a 99.999\% percentile-based range calibration to remove outliers. The quantized range is always defined by the range representable by a signed integer at a given bit-precision.

We vary quantization \emph{granularity}, \emph{symmetry}, and \emph{calibration mode}: 
\emph{Granularity} specifies whether the quantization operation is to be performed on the entire tensor (\textit{per-tensor}) or individually on different channels of the tensor (\textit{per-channel}). In the case of the S4D, $H$ independent \acp{SSM} are stacked in parallel, as observed in \cref{algo:S4D_summary}. Formally, the weights, states and activations of each head can be aggregated into respective tensors with an $H$ dimension. Per-channel or per-tensor quantization can then be applied to this tensor. In this case of the SSM, we emphasize that per-channel quantization of the state-space matrices is performed over the $H$ dimension by referring to it as \textit{per-head}. 
\emph{Symmetry} describes the assumptions on the full-precision range. The full-precision range can be taken to be symmetric around zero (\textit{symmetric}), which is then mapped to the quantized range through a linear scaling, or as an arbitrary range that requires an affine mapping to the quantized range (\textit{asymmetric}). 
The latter is more accurate, but also more computationally and memory-expensive as it complicates quantized arithmetic and requires storing the zero-point of the mapping. 
Finally, the \emph{calibration mode} determines whether activations and other runtime-computed values (such as state) are quantized \textit{statically}, by pre-calibrating to obtain range statistics fixed during deployment, or \textit{dynamically} by adjusting the statistics each time new inputs are presented. 
The latter mode is impractical for deployment in edge processing cases, and we generally implement static quantization unless specified otherwise.

For \ac{QAT}, we perform 10 additional epochs \textit{in recurrent mode} on the pre-trained quantized model with a learning rate of $1\mathrm{e}{-4}$ (this is also known as quantization-aware fine tuning (QAFT)). 
As is customary, we utilize the straight-through gradient estimator (STE) \cite{bengio2013estimating} to allow backpropagation of errors through non-differentiable operations. 
To prevent exploding gradients in the recurrence, we further constrain the gradients to the range $[-1000, 1000]$. 
All experiments use asymmetric, per-head, and static quantization schemes.

A crucial element for consideration during \ac{QAT} is the parameterization of the state-space matrices, which is important for model stability during training. Parametrization refers to the functional form utilized to express a parameter; in the case of \acp{SSM}, it is used to condition the training of $\bar{A}$ and avoid instability in the state.
In the original S4D model, the state-space matrices (and $\Delta$) are trained in continuous-time, where $A$ is further parameterized by independently training the negative log of the real component and the imaginary component. 
In our case, one can directly train the quantized discrete state-space matrices or continue to train the continuous-time state-space matrices in the original parametrization.
We refer to these parameterizations as \textit{Discrete} and \textit{Continuous} (or \textit{Original}), respectively. Additionally, we experiment with \textit{Frozen $\bar{A}$}, meaning that we train the discrete state-space matrices, but apply no updates to $\bar{A}$ during \ac{QAT}. 

\begin{figure}[t!]
    \centering
    \includegraphics[width=0.99\linewidth]{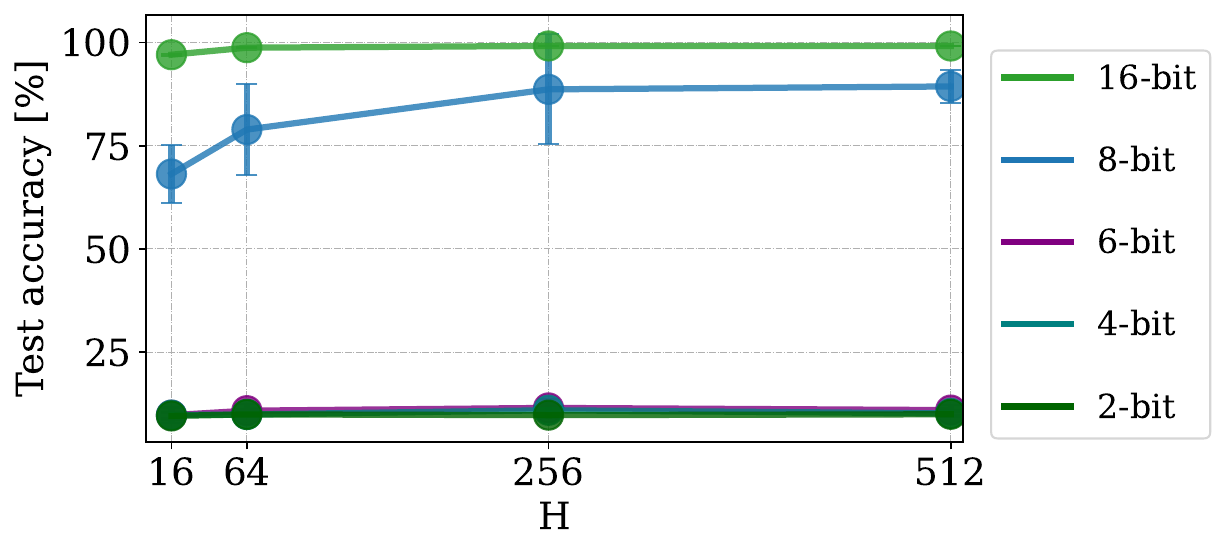}
    \caption{Effect of scaling the model width on \ac{PTQ} performance. H is the number of heads in the SSM layer.}
    \label{fig:PTQ-scaling}
\end{figure}

\begin{table}[ht!]
	\centering
        \caption{State transition matrix parameterization schemes for \ac{QAT}. ``Discrete'' trains $\bar{A}$ directly, ``Continuous'' uses the original S4D parameterization and trains the continuous-time $A$.}
	\begin{tabular}{p{1.5cm}|p{3cm}|p{3.cm}}
		Name & Trainable Parameters & Parameterization of $\bar{A}$ \\
		\hline
		&\\[-0.5em]
		Discrete & $\bar{A}, \bar{B}, \bar{C}, \bar{D}$ & $\bar{A}$ = $\bar{A}$ \\ 
		&\\
		Continuous \newline (original) & $A_{Re}$, $A_{Im}$, $B$, $C$, $D$, $\Delta$ & $A = -e^{A_{Re}} + iA_{Im}$ \newline  $\bar{A} = \text{Disc}(\bar{A}, \Delta)$  \\
		&\\
		\hline
	\end{tabular}
    \label{tab:params}
\end{table}

\begin{figure*}[t!]
    \centering

    \begin{subfigure}[b]{0.3\textwidth}
        \centering
        \includegraphics[width=\textwidth]{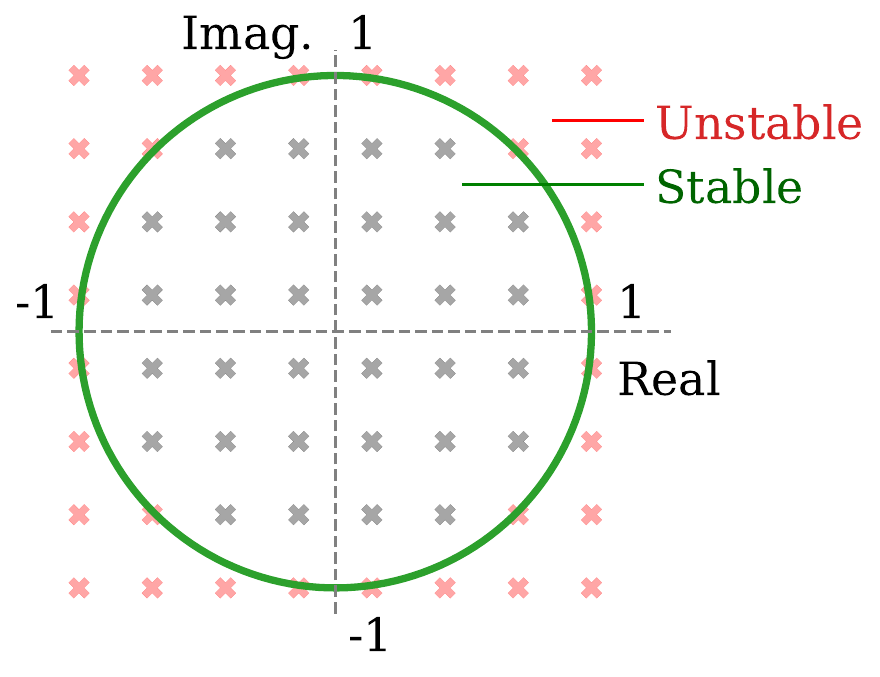}
        \caption{}
        \label{fig:quantization_complex}
    \end{subfigure}
    \hspace{1cm}
    \begin{subfigure}[b]{0.41\textwidth}
        \centering
        \includegraphics[width=\textwidth]{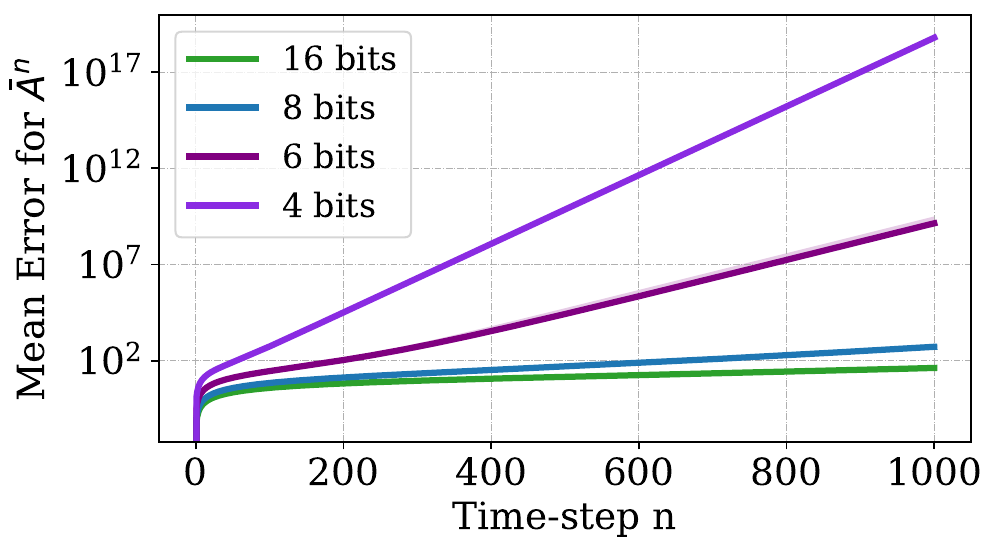}
        \caption{}
        \label{fig:F3_b}
    \end{subfigure}
    
    \caption{Quantization of complex-valued parameters. (a) Real and Imaginary parts are quantized independently, here with 3-bit precision, with quantized values represented with crosses. The red area around the unit circuit is unstable, while crosses landing within the circle are stable. (b) Mean Euclidean distance (mean error $||A^n_{fp} - A^n_q||_2$) between the full-precision $A_{fp}$ and the quantized $A_q$ for different quantization bit precisions, and different time steps (powers of $n$).}
    \label{fig:F3}
\end{figure*}

To summarize, we always apply the quantization function $Q(\cdot)$ after the discretization function $\text{Disc}(\cdot)$. In our experiments, we use ZOH discretization. For the continuous parametrization of $\bar{A}$, the quantized and discretized $\bar{A}_q$ is given as:
\begin{equation}
    \bar{A}_q = Q(\text{Disc}( \bar{A}) ) = Q(e^{A\Delta})
\end{equation}
For all experiments, we apply \textit{fake-quantization} \cite{jacob2018quantization} to the S4D parameters, which simulates the effects of quantization through rounding and clipping. The quantization function $Q(\cdot)$ rounds and clips its argument to the full-precision values representable by a given bit-precision.

All \ac{PTQ} and \ac{QAT} experiments are initialized with 5 distinct seeds. When applicable, standard deviations of the results of different seeds are shown as error bars.

\section{Results on post-training quantization}
In this section, we analyze the effect of quantizing the pre-trained full-precision S4D models with a thorough ablation study to identify the sensitivity of different model components to quantization parameters and bit-precision.
We perform all the experiments in this section on the smallest S4D model with 16 heads (S4D-16h), as its small size exacerbates parameter sensitivity to quantization.
Unless specified otherwise, we quantize weights and activations with a static, per-head, and asymmetric quantization scheme.

\paragraph{SSM weights}
We first analyze the effect on model performance of independently quantizing each state-space matrix ($\bar{A}$, $\bar{B}$, $\bar{C}$, and $\bar{D}$). All other parameters and activations remain in full-precision. As shown in ~\cref{fig:ptq-a}, the state transition matrix $\bar{A}$ is the most critical in determining the performance of the SSM, and quantizing it to 8-bit precision already incurs a heavy performance penalty. Quantizing $\bar{B}$, $\bar{C}$, and $\bar{D}$ has little effect on performance down to 4-bit precision. In general, $\bar{C}$ is slightly more sensitive than $\bar{B}$ and $\bar{D}$.

\paragraph{State and activations}
Next, we focus on the runtime-computed values, including all activations, in the S4D architecture, with emphasis on the \ac{SSM} state. For this analysis, we evaluate both \textit{static} and \textit{dynamic} quantization modes for the either the state with all other runtime-computed values in full precision, or all runtime-computed values with only the state in full precision. Recall that the \textit{static} mode is easier to implement in hardware, while the \textit{dynamic} mode is expected to be more precise.
Similar to the $\bar{A}$ parameter, the state $x$ precision is a crucially important factor in post-quantization performance (\cref{fig:ptq-b}). In fact, 8-bit quantization suffices to disrupt performance. Moreover, \textit{dynamic} quantization mode performs much better than the \textit{static} mode at the boundary precision before model failure, but it is impractical in resource-constrained hardware at the edge. The other runtime-computed values in ~\cref{fig:ptq-b} can be more aggressively quantized, and 6 bits are enough to maintain high accuracy.

\paragraph{Quantization granularity}
We further explore quantization granularity. We analyze the effect of the two quantization modes in~\cref{fig:ptq-c}, where all parameters are at quantized at the same bit-precision. In the \textit{per-tensor} case, classification accuracy drops to chance level already at 8-bit precision. As expected, \textit{per-head} quantization yields better performance than \textit{per-tensor}, as it often yields higher-resolution quantization ranges customized for each \ac{SSM} head. This is mainly visible at the 8-bit boundary of model failure. 

\paragraph{Quantization symmetry}
We apply either \textit{symmetric} or \textit{asymmetric} quantization to all parameters and activation quantized at the same bit precision. As expected, the model's performance significantly drops at 8-bit precision. \textit{Asymmetric} quantization yields slightly better classification accuracy than the \textit{symmetric} case, but the difference is only noticeable at 8-bit precision.

\paragraph{Scaling up to wider models}
Our previous experiments are performed with the smallest S4D size, where the effects of quantization are most easily observable. 
We observed that the state transition matrix $\bar{A}$ and state \textit{x}, under \ac{PTQ}, require a minimum precision of 8 bits.
Here, we investigate whether this boundary holds when scaling up the width of the SSM layer by increasing the number of heads. We increase the number of heads from 16 to 512, thus increasing the model size from about 4k to 630k, while quantizing all weights and activations at a precision varying from 2 to 16 bits. Results are shown in~\cref{fig:PTQ-scaling}. Even for larger models, bit-precisions lower than 8-bit completely hinders the model performance.
Only at the 8-bit critical point, before catastrophic model failure, does the model size positively correlate with classification accuracy. Nevertheless, scaling model width does not permit the 8-bit model to approach the performance achieved by the 16-bit model.

In summary, larger models generally do not permit more aggressive PTQ-quantization, except specifically at the critical bit-precision. When using only \ac{PTQ}, the catastrophic effects of poor quantization schemes persist for larger models. Classification accuracy remains limited by the requirement of high bit-precision in the SSM layer.

\begin{figure*}[t!]
    \centering
    \begin{subfigure}[b]{0.38\textwidth}
        \centering
        \includegraphics[width=\textwidth]{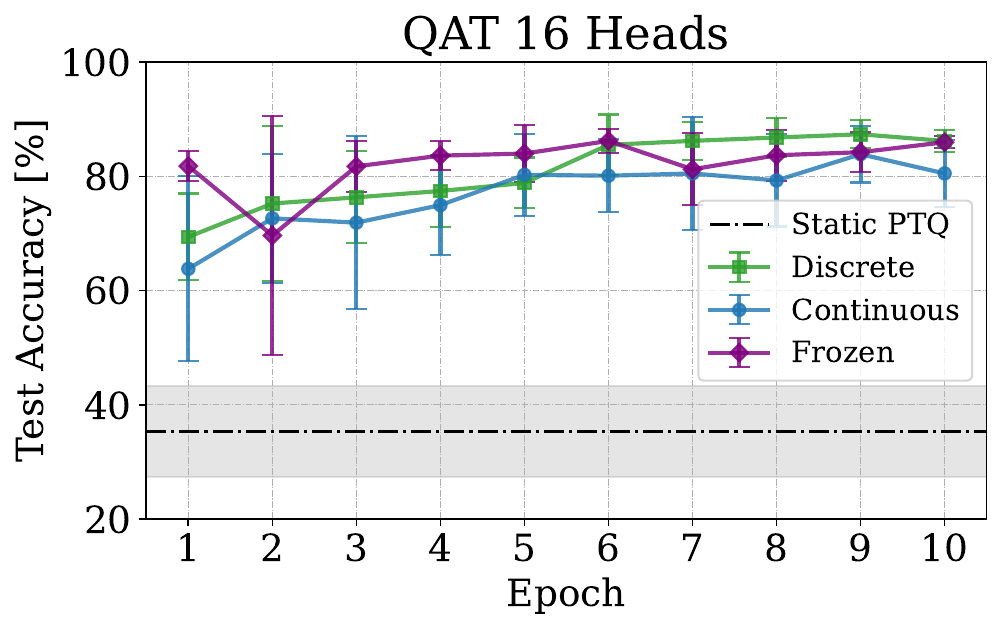}
        \caption{}
        \label{fig:F4_a}
    \end{subfigure}
    \hspace{1cm}
    \begin{subfigure}[b]{0.38\textwidth}
        \centering
        \includegraphics[width=\textwidth]{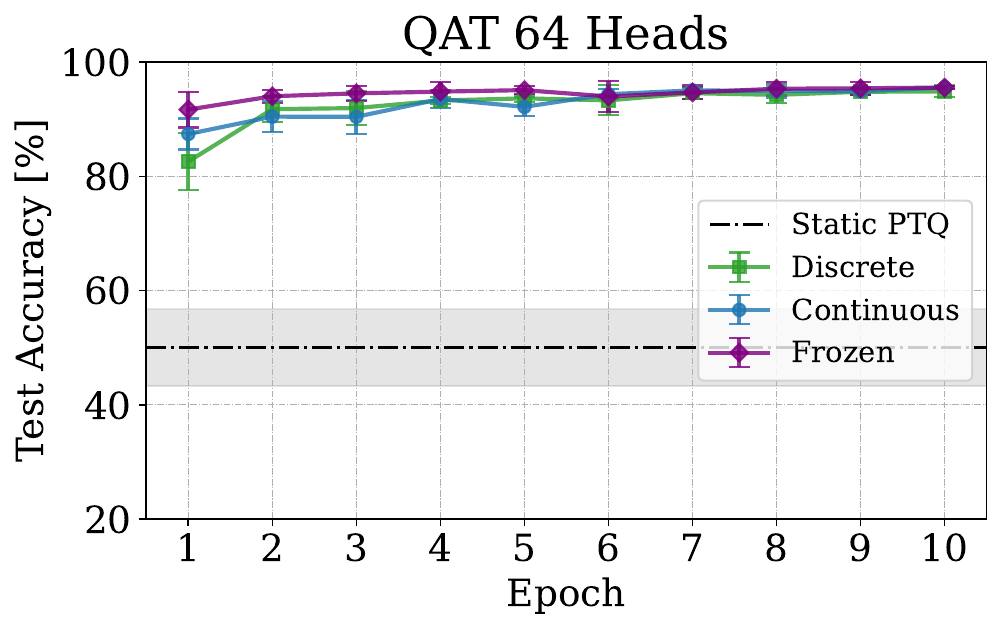}
        \caption{}
        \label{fig:F4_b}
    \end{subfigure}

    \vspace{0.cm} 

    \begin{subfigure}[b]{0.95\textwidth}
        \centering
        \includegraphics[width=\textwidth]{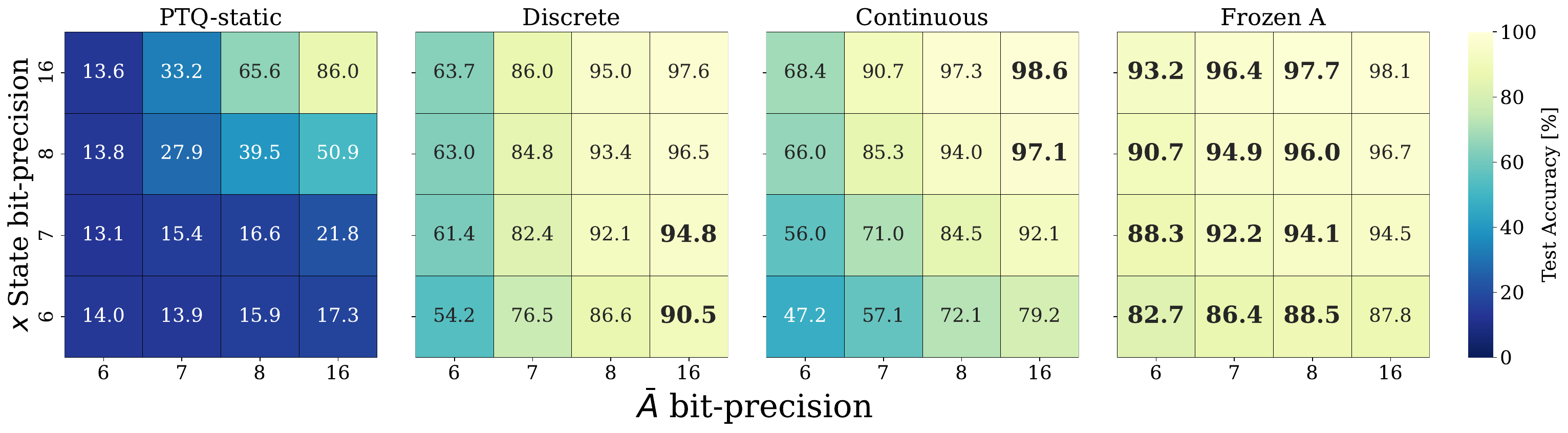}
        \caption{}
        \label{fig:F4_c}
    \end{subfigure}
    
    \caption{\Ac{QAT} on S4D model with 16 (a) and 64 (b) heads, with all parameters and activations quantized at 8 bits, using asymmetric and static schemes. Experiments in (a) and (b) are performed on 5 different seeds, and standard deviations are shown with error bars. (c) Performance of \ac{QAT} for S4D with 64 heads for different bit-precision of $\bar{A}$ and \textit{x}.}
    \label{fig:F4_QAT}
\end{figure*}

\section{Stability of the state transition matrix state}
The state transition matrix $\bar{A}$ and state \textit{x} are extremely sensitive to quantization. In this section, we analyze the cause of this sensitivity. Recall that $\bar{A}$ is expressed in the complex domain, and we separate its real and imaginary components before quantization. However, we always quantize both components with the same scheme, mapping the full-precision range to the quantized range with the same scaling and offset. This is depicted in \cref{fig:quantization_complex} where the quantized space (a uniform Cartesian grid) is marked as crosses in the complex plane.

The LTI system (\cref{eq:LTI_eqns}) can only be stable when the real components of all eigenvalues of $A$ are strictly negative. In the discrete system, the eigenvalues of $\bar{A}$ must have magnitudes strictly less than 1.
In the case of the S4D model, all elements of the $\bar{A}$, which is already diagonal, must land within the unit circle (in green in \cref{fig:quantization_complex}).
However, even if all full-precision values are within the unit circle, the quantized space nevertheless allows values outside the unit circle (red crosses in~\cref{fig:quantization_complex}), resulting in potential instability. With low bit-resolution, the number of stable quantized values close to the unit circle reduces, and the entries in $\bar{A}$ are quantized to the nearest representable value. If the nearest available quantized value for an entry of $\bar{A}$ falls outside the unit circle, then the SSM will be unstable.
Errors in the quantization of $\bar{A}$ are particularly critical, as the state \textit{x} is repeatedly multiplied by $\bar{A}$ within the recursion. Therefore, errors due to the quantization of $\bar{A}$ are amplified by the power of $n$, where $n$ is the time step, causing \textit{x} to explode.
This is shown in~\cref{fig:F3}b, which approximates quantization error by showing the mean absolute error between the quantized and full-precision $\bar{A}^n$.

An additional advantage of operating the model in recurrent mode is that the state can be constrained after each time step. Rather than explicitly constraining the eigenvalues of $\bar{A}$ within the unit circle, we found that simply clipping $x$ within a bounded range at each time step offers a simple, robust, and general solution to the instability problem during both \ac{PTQ} and \ac{QAT}.

\section{Results on quantization-aware training}

We perform \ac{QAT} to improve the performance of the quantized S4D models.
Starting from the pre-trained quantized model, we train both the small (16 heads) and medium (64 heads) quantized models in recurrent mode. 
We analyze three distinct parameterizations of $\bar{A}$: ``Discrete'', ``Continuous'', and ``Frozen $\bar{A}$'', discussed in \cref{sec:method}.
In all cases, the classification accuracy greatly increases after just one \ac{QAT} epoch (\cref{fig:F4_QAT}(a), (b)), and it reaches convergence within 10 epochs. On the smaller model size (16 heads), the ``Continuous" S4D parametrization underperforms compared to the ``Frozen $\bar{A}$" and ``Discrete" ones. On the 64-head model, ``Frozen $\bar{A}$" parameterization converges faster and reaches slightly higher accuracy.

\paragraph{Heterogeneous quantization scheme}
Our \ac{PTQ} analysis showed that different weights and activations in the S4D model display markedly different sensitivities to quantization.
Consequently, we introduce a heterogeneous scheme that quantizes parameters and states at different bit-precisions. Such a scheme is necessary to optimize memory usage for hardware deployment.

Outside the SSM layer, we quantize all parameters to 4 bits. Within the SSM layer, we also apply 4-bit quantization to the $\bar{B}$, $\bar{C}$, and $\bar{D}$ state-space matrices. All activations, except the state, are quantized to 6 bits.
Meanwhile, we investigate how far one can reduce the bit-precision for $\bar{A}$ and \textit{x}. 

\Cref{fig:F4_QAT}c shows final test accuracies after \ac{QAT} for the three proposed parameterizations of $\bar{A}$ and compares them with the accuracies after \ac{PTQ}. At high-precision $\bar{A}$ and \textit{x}, continuous parameterization performs best, while at high-precision $\bar{A}$ and low-precision $x$, directly training the state space matrices with discrete parameterization yields the best performance. Most importantly, the frozen $\bar{A}$ parameterization consistently outperforms the other two parameterizations on lower $\bar{A}$ precisions, which is the most hardware-relevant domain.

While this is surprising, as the model trains fewer parameters, this parameterization could be a pragmatic choice in cases where low bit-precision quantization does not allow fine-tuning of the state transition matrix $\bar{A}$. The performance of this parameterization during \ac{QAT} suggests that the best choice of quantized $\bar{A}$ at low precisions is the one that is closest to its floating-point value.

To summarize, in \cref{tab:qat-results} we report the results obtained during the \ac{QAT} analysis, including the model's classification accuracy and the memory savings compared to the full-precision baselines. Note that the 64-head model with heterogeneous quantization and 8-bit precision in the state transition matrix $\bar{A}$ and state \textit{x} initially perform at around 40\% after \ac{PTQ}, but after \ac{QAT} attains 95.99\% classification accuracy. This performance is achieved while reducing the full-precision model's memory footprint by 84.05\% , equal to a 6x compression.

\begin{table}[h]
	\centering
        \caption{Summary of memory footprint savings from \ac{QAT} results. Accuracy and standard deviation of the best-performing parameterization and quantization scheme on sMNIST. ``W[X]A[Y]" denotes bit-precision of X for weights, Y for activations. $\bar{A}$ and $x$, if quantized differently from other weights and activations, are also denoted explicitly with their precisions.}
    \resizebox{\linewidth}{!}{%
	\begin{tabular}{c|c|c|c}
		\textbf{Model} & \textbf{Quantization} &\textbf{Accuracy [\%]} & \textbf{Mem Savings [\%]} \\
		\hline
		\multirow{6}{*}{S4D-16} & FP & 97.20 $\pm$ 0.25 & - \\
		& PTQ W4A6 $\bar{A}$8$x$8 & 29.36 $\pm$ 14.58 & 82.96 \\
            & W8A8 & 92.64 $\pm$ 0.62 & 76.55 \\
		& W4A6, $\bar{A}$16$x$16 & 96.59 $\pm$ 0.30 & 73.54 \\ 
		& W4A6, $\bar{A}$8$x$8 & 88.39 $\pm$ 1.40 & 82.96 \\
		& W4A6, $\bar{A}$6$x$6 & 57.04 $\pm$ 3.13 & 85.31 \\
		\hline
		\multirow{6}{*}{S4D-64} & FP & 98.79 $\pm$ 0.12 & - \\
            & PTQ W4A6 $\bar{A}$8$x$8 & \textcolor{orange}{39.46} $\pm$ 14.98 & \textcolor{OliveGreen}{\textbf{84.05}} \\
		& W8A8 & 97.66 $\pm$ 0.30 & 76.29 \\
		& W4A6, $\bar{A}$16$x$16 & 98.62 $\pm$ 0.08 & 76.70 \\ 
		& W4A6, $\bar{A}$8$x$8 & \textcolor{OliveGreen}{\textbf{95.99}} $\pm$ 0.56 & \textcolor{OliveGreen}{\textbf{84.05}} \\
		& W4A6, $\bar{A}$6$x$6 & 84.21 $\pm$ 2.53 & 85.89 \\
		\hline		
	\end{tabular}
    }
        \label{tab:qat-results}
    
\end{table}

\section{Conclusion}
\acp{SSM} are formidable neural network architectures capable of processing long-range sequences with excellent parameter efficiency. In this work, we analyze the compression of the memory footprint of the S4D \cite{meyer2024diagonal} \ac{SSM} towards hardware deployment with constrained resources at the edge. In the context of a common sequence classification task (sMNIST), we analyze the criticality of each parameter in the \ac{SSM} network by applying \ac{PTQ} to the \ac{SSM} parameters at different bit-widths. We show that the precision of the state matrix $\bar{A}$ and state $\textit{x}$ are critical to performance. When using only \ac{PTQ}, these values require at least 8-bit precision to perform well in small scale models, which remains true even with larger network layers. For this reason, we perform \ac{QAT} to reduce the bit-width requirement for the \ac{SSM} parameters. We demonstrate that it is indeed possible to lower the precision of $\bar{A}$ and $\textit{x}$ below 8 bits without catastrophic model failure, but 8-bit precision represents a reasonable compromise between memory footprint compression and performance. We develop a heterogeneous quantization scheme whereby different \ac{SSM} parameters are quantized at different bit-precisions. To better optimize memory usage on hardware, our quantization scheme exploits the fact that different \ac{SSM} parameters have different sensitivities to quantization.
Combining \ac{QAT} with the heterogeneous quantization technique, we recover a classification accuracy of 96\% with \ac{QAT} compared to 40\% after \ac{PTQ}, while reducing the full-precision model's memory footprint by a 6x factor. This scheme strikes a balance between high performance, low memory footprint and computational complexity, optimizing the S4D model for execution on an edge processor.

\subsection{Outlook}
This work paves the way towards the deployment of efficient yet powerful state-space models in edge computing cases. Different SSM variants have been developed, and it remains to be seen which is the most adapted to edge machine learning, presenting the best compromise between memory footprint, number of operations, and performance in relevant tasks. The S4D \cite{gu2022parameterization} model offers efficient execution thanks to the diagonal structure in the $A$ matrix. The same holds for the S5 \cite{smith2022simplified} and S7 \cite{soydan2024s7} models, while the S6 alternative also simplifies the non-linearity to a binarized one, potentially helping reduce the number of operations at inference. The S4 \cite{gu2111efficiently} model and Mamba \cite{gu2023mamba} are instead interesting options for natural language processing tasks.

The heterogeneous quantization scheme detailed in our approach incurs some computational overhead to align the integer ranges. Future work will address the hardware cost of on-device dequantization and quantization operations.

Furthermore, such quantized models remain to be tested on relevant tasks in the context of edge AI. An example is the Google Speech Command \cite{warden2018speech} benchmark tasks, a keyword spotting task. Other examples are bio-medical signal processing (ECG \cite{kim2023tinyml}, EEG, and EMG \cite{donati2023long}) or the processing of sensory signals from robots and drones.

\section*{Acknowledgment}
We acknowledge the entire EIS lab for their support throughout the development of this project. The presented work has received funding from the Swiss National Science Foundation Starting Grant Project UNITE (TMSGI2-211461).


\end{document}